\documentclass[letterpaper]{article} % DO NOT CHANGE THIS
\usepackage{aaai25}  % DO NOT CHANGE THIS
\usepackage{times}  % DO NOT CHANGE THIS
\usepackage{helvet}  % DO NOT CHANGE THIS
\usepackage{courier}  % DO NOT CHANGE THIS
\usepackage[hyphens]{url}  % DO NOT CHANGE THIS
\usepackage{graphicx} % DO NOT CHANGE THIS
\usepackage{amsmath}
\usepackage{multirow}
\usepackage{natbib}  % DO NOT CHANGE THIS AND DO NOT ADD ANY OPTIONS TO IT
\usepackage{caption} % DO NOT CHANGE THIS AND DO NOT ADD ANY OPTIONS TO IT
\usepackage{algorithm}
\usepackage{algorithmic}

\usepackage{newfloat}
\usepackage{listings}
\DeclareCaptionStyle{ruled}{labelfont=normalfont,labelsep=colon,strut=off} % DO NOT CHANGE THIS
\floatstyle{ruled}
\newfloat{listing}{tb}{lst}{}
\floatname{listing}{Listing}
\usepackage{bibentry}
\begin{document}

\title{Equitable Skin Disease Prediction\\ Using Transfer Learning and Domain Adaptation}

\author{
    Sajib Acharjee Dip\textsuperscript{\rm 1$\dagger$}, 
    Kazi Hasan Ibn Arif\textsuperscript{\rm 1$\dagger$}, 
    Uddip Acharjee Shuvo\textsuperscript{\rm 2}, 
    Ishtiaque Ahmed Khan\textsuperscript{\rm 1}, 
    Na Meng\textsuperscript{\rm 1*}
}

\affiliations{
    \textsuperscript{\rm 1}Department of Computer Science, Virginia Tech\\
    \textsuperscript{\rm 2}Institute of Information Technology, Dhaka University\\
    \texttt{\{sajibacharjeedip, hasanarif, ishtiaqueahmedk\}@vt.edu}, 
    \texttt{uddip.acharjee.shuvo@gmail.com}, 
    \texttt{nm8247@vt.edu}
}

\maketitle

\begin{abstract}
In the realm of dermatology, the complexity of diagnosing skin conditions manually necessitates the expertise of dermatologists. Accurate identification of various skin ailments, ranging from cancer to inflammatory diseases, is paramount. However, existing artificial intelligence (AI) models in dermatology face challenges, particularly in accurately diagnosing diseases across diverse skin tones, with a notable performance gap in darker skin. Additionally, the scarcity of publicly available, unbiased datasets hampers the development of inclusive AI diagnostic tools. To tackle the challenges in accurately predicting skin conditions across diverse skin tones, we employ a transfer-learning approach that capitalizes on the rich, transferable knowledge from various image domains. Our method integrates multiple pre-trained models from a wide range of sources, including general and specific medical images, to improve the robustness and inclusiveness of the skin condition predictions. We rigorously evaluated the effectiveness of these models using the Diverse Dermatology Images (DDI) dataset, which uniquely encompasses both underrepresented and common skin tones, making it an ideal benchmark for assessing our approach. Among all methods, Med-ViT emerged as the top performer due to its comprehensive feature representation learned from diverse image sources. To further enhance performance, we conducted domain adaptation using additional skin image datasets such as HAM10000. This adaptation significantly improved model performance across all models.

% Our approach aims to leverage transfer learning techniques to fine-tune foundation models pre-trained on diverse domains, thereby enhancing the accuracy and inclusivity of skin disease prediction. This project report outlines the evolving landscape of dermatological AI, underscores the importance of diverse datasets. It also elucidates how our refined methodology bridges existing research gaps, promising more inclusive and accurate skin disease diagnosis. Code:https://github.com/Sajib-006/diverse-dermatology-prediction

\end{abstract}

% Uncomment the following to link to your code, datasets, an extended version or similar.
%
% \begin{links}
%     \link{Code}{https://aaai.org/example/code}
%     \link{Datasets}{https://aaai.org/example/datasets}
%     \link{Extended version}{https://aaai.org/example/extended-version}
% \end{links}

\section{Introduction}
Skin diseases encompass a wide spectrum of conditions, and some pose significant health risks if not identified and treated promptly. The diagnosis of these diseases is predominantly a manual process performed by dermatologists through visual inspection and clinical judgment. As the prevalence of skin diseases increases worldwide, the need for an efficient and accurate diagnosis becomes increasingly pressing. AI has emerged as a promising solution to assist in the triage and preliminary identification of skin conditions, potentially leading to early intervention and better patient outcomes \cite{gondocs2024ai, du2020ai, gomolin2020artificial, hogarty2020artificial}.

However, the effectiveness of AI in dermatology is currently hindered by two main challenges: the limited performance of existing models on diverse skin tones and the absence of comprehensive and unbiased datasets that reflect the full spectrum of skin diseases across different ethnicities. The introduction of the Diverse Dermatology Images (DDI) dataset is a commendable step toward rectifying the latter issue, although its small size presents challenges for conventional deep learning applications \cite{daneshjou2022disparities}.

In the realm of medical AI, prior research has underscored the effectiveness of transfer learning, particularly when confronted with limited dataset sizes for training deep learning models from scratch \cite{alzubaidi2021novel, dip2024patholm, yu2022transfer, bi2024ai, paul2022candidates}. In particular, Vision transformer \cite{dosovitskiy2020image} based foundation models like RETFound, initially trained on extensive retinal imaging datasets such as ImageNet-1k \cite{5206848} and MEH-MIDAS, have shown promise in capturing intricate domain-specific features transferable to diverse medical imaging tasks \cite{zhou2023foundation}.

Our contribution extends beyond previous efforts by adapting these domain-specific foundation models to the domain of skin disease classification. We propose a novel approach that takes advantage of transfer learning to bridge the gap between retinal and dermatological imaging. We take advantage of the features learned from one medical domain to enrich understanding in another. Specifically, we introduce RETFound alongside other pre-trained models such as MedViT \cite{manzari2023medvit} and YOLOv8-Chest (pre-trained on chest images) \cite{reis2023real}, highlighting their diverse capabilities in capturing relevant medical features.

Our methodology entails benchmarking these models, including YOLOv8, YOLOv8-Chest, MedViT, and RETFound, to assess their performance on skin disease classification tasks. By fine-tuning these models using the Diverse Dermatology Images (DDI) dataset, we aim not only to overcome the constraints imposed by the size of the dataset but also to harness the potential of learned medical imaging features from related domains. Furthermore, we incorporate domain adaptation techniques, using larger related data sets like HAM10000 \cite{tschandl2018ham10000}, to further enhance model performance, particularly in diagnosing skin diseases in diverse skin tones.

This comprehensive benchmarking of pre-trained models in diverse unbiased skin image prediction represents a significant contribution to the field. By showcasing how domain adaptation and transfer learning techniques can leverage pre-trained knowledge to improve performance, especially in underrepresented skin tone scenarios, our approach promises to advance the development of more equitable AI tools in dermatology, ensuring inclusivity and accuracy across diverse patient populations.

\section{Related Work}
Prior studies have validated the efficacy of machine learning (ML) and deep learning (DL) in the classification and diagnosis of dermatological conditions, achieving levels of performance comparable to or exceeding that of board-certified dermatologists in cases of skin cancer \cite{brinker2019convolutional,esteva2017correction}, eczema \cite{de2015design}, psoriasis \cite{shrivastava2016computer}, and onychomycosis \cite{han2018deep}. In particular, Emam et al. \cite{emam2020predicting} reported an AUC of up to 0.95 for the discontinuation of biological treatments using a variety of models, including deep learning techniques. Similarly, Wang et al. \cite{wang2019assessment} and Roffman et al. \cite{roffman2018predicting} focused on predicting non-melanoma skin cancer with AUCs of 0.89 and 0.81, respectively. The work of Khozeimeh et al. \cite{khozeimeh2017expert} presented a distinction in the response to wart treatment methods between cryotherapy and immunotherapy, with respective accuracies of 80\% and 98\%. Furthermore, Tan et al. \cite{tan2017practical} investigated the complexity of reconstructive surgery after excision of periocular basal cell carcinoma, applying Bayesian and other methodologies to achieve AUC values greater than 0.83. Each of these investigations used data sets that encompass 7 to 20 clinically relevant patient characteristics, underscoring the importance of comprehensive data for model training and validation. Egorov et al. \cite{egorov2009differentiation} evaluated three advanced models: ModelDerm \cite{han2020augmented}, DeepDerm \cite{esteva2017dermatologist}, and HAM10000 \cite{tschandl2018ham10000}, which showed commendable results in the datasets on which they were trained, but experienced a decrease in performance when applied to the DDI \cite{daneshjou2022disparities}. Therefore, we can say, existing dermatological diagnostic algorithms lacks robustness and generalizability. Our research seeks to address this gap.

\section{Methods and Materials}
\subsection{Dataset Collection }
In our methodological framework, the collection of data sets is significant, as it underpins the training of our AI model. The primary dataset utilized in this study is an assembly of skin disease images with a focus on inclusive skin tone representation. These images were meticulously curated from pathology reports archived at the Stanford Clinic over a decade from 2010 to 2020. To ensure the reliability and clinical applicability of the dataset, each image has been annotated by a duo of board-certified dermatologists, providing a substantial foundation for the subsequent AI-driven analysis. The data set embraces the Fitzpatrick Skin Type (FST) classification system, a globally recognized schema for categorizing human skin tones. This stratification allows for a detailed and nuanced approach to the representation of diverse skin types within our dataset. In total, the data set comprises 656 images depicting conditions of 570 unique patients. These images are distributed across the FST spectrum as follows: 208 images from FST categories I-II, including 159 benign and 49 malignant cases; 241 images from FST III-IV, encompassing 167 benign and 74 malignant cases; and 159 images from FST V-VI, with 159 benign and 48 malignant cases.

To augment our primary dataset and enhance the robustness of our transfer learning approach, we have incorporated additional datasets renowned for their extensive collection of images of skin disease. DeepDerm provides a vast repository with 129,450 images, and Ham10000 complements this with an additional 10,015 images. These datasets serve as a foundation for the initial adaptation phase of our pre-trained model, enabling it to acclimate to the domain of dermatological imagery before fine-tuning with the more focused but less voluminous DDI dataset. The strategic amalgamation of these datasets is designed to foster a comprehensive learning environment that allows the extraction of generalizable features, which are then refined to discern subtle nuances across diverse skin types.

\subsection{Transfer learning}
In our transfer learning approach, we begin with a pre-trained model on retinal images, using it as a foundation to adapt to dermatological tasks with the DeepDerm and HAM10000 datasets. We then finetune the model's weights on the DDI dataset to refine its ability for skin disease classification, ensuring specificity and accuracy in our predictions.

\subsection{Model Selection}
\subsubsection{Skin Image Pre-trained Models}
We select DeepDerm \cite{esteva2017dermatologist} and HAM10000 \cite{tschandl2018ham10000} as open-source pre-trained models trained on skin images for dermatology applications. While selecting models, we focuses on their adaptability for skin cancer classification and reliable accuracy. DeepDerm is trained end-to-end from images and disease labels and performs comparable to board-certified dermatologists in the classification of skin lesions. It shows potential for enhanced diagnosis using deep convolutional neural networks (CNNs) and a dataset of 129,450 clinical images. Furthermore, HAM10000 overcomes the challenge of diversity in dermatoscopic image datasets with 10,015 images, facilitating machine learning research and comparisons with human experts in diagnosing pigmented skin lesions, with more than 50\% of the lesions confirmed by pathology.

\subsubsection{Other medical domain Pre-trained Model}
In selecting our model, we focus on the domain-specific RETFound, pre-trained on ImageNet-1k and MEH-MIDAS datasets, as depicted in Figure \ref{fig:model}. For benchmarking, we use the generalist Vision Transformer (ViT) trained on ImageNet-21k to contrast its performance with our specialized approach.

\subsubsection{General Medical Image Pre-trained Model}
There exist many general-purpose Medical Imaging models. These models are commonly used in various downstream tasks through fine-tuning. These models are trained on diverse types of medical images representing different body parts and organs. We choose MedViT \cite{manzari2023medvit} as our pre-trained model. It introduces a hybrid CNN-Transformer model that merges the CNN's locality with the vision Transformer's global connectivity. MedViT stands out for its focus on learning smoother decision boundaries to increase resilience against adversarial attacks. This is achieved by augmenting shape information within the high-level feature space. In particular, this model demonstrates high robustness and generalization capabilities while managing to reduce computational complexity. Its performance sets a new benchmark in medical image analysis. For our specific classification task, we adapted and fine-tuned the MedViT model using skin datasets.

\subsubsection{General Image Pre-trained Model}
In our exploration of the transfer learning approach, we extend our scope to include general purpose vision models. YOLOv8 \cite{redmon2016you} is known as a leading contender in this category. The model achieves state-of-the-earth results in various real-time object detection and image segmentation capabilities. First introduced in 2015, YOLO (You Only Look Once) quickly gained acclaim for its exceptional speed and accuracy. We use the latest version of the model. To fit YOLO as an effective classifier model between benign and malignant cancers images, we replace it's final layers with a classification layer and fine tune the whole model with our curated skin dataset.

\begin{figure*}[h!] % h! forces the figure to be placed here
\centering
\includegraphics[width=0.95\textwidth]{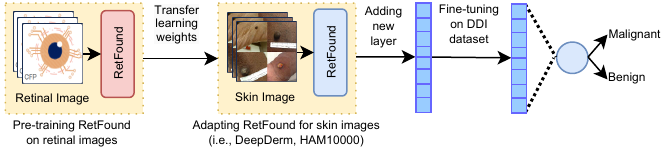} % Adjust the size of the image

\caption{\textbf{Model Architecture with Pre-training and Transfer Learning.} The diagram illustrates the two-phase model development process. The upper section depicts pre-training on extensive datasets, enhancing foundational knowledge across varied domains. The lower section demonstrates fine-tuning on skin disease datasets, specifically adapting the model for binary classification of skin diseases. The RETFOUND model, exemplified here, undergoes enhancement through transfer learning by incorporating new layers. This approach culminates in a refined classification model adept at predicting skin diseases based on learned patterns and features.
}
\label{fig:model}
\end{figure*}
% \section{Implementation}
\subsection{Data Preprocessing}

In the data processing phase of the study, we meticulously curated the Diverse Dermatology Images (DDI) dataset to ensure a comprehensive and balanced evaluation of all models involved. This dataset, sourced from Stanford Clinic Pathology reports spanning from 2010 to 2020, comprises images labeled by two boards of certified dermatologists, providing a robust foundation for our research.

One of the unique characteristics of the DDI dataset is its representation of diverse skin tones, classified according to the Fitzpatrick Skin Type (FST) scheme. We categorized the dataset into three distinct skin tone groups:

\begin{itemize}
    \item Dark Skin Tone (FST I-II): This group encompasses individuals with darker skin tones, represented by FST categories I and II.
    \item Medium Skin Tone (FST III-IV): Individuals with medium skin tones fall under FST categories III and IV.
    \item White Skin Tone (FST V-VI): FST categories V and VI represent individuals with lighter skin tones.
\end{itemize}

Each skin tone group contains images labeled with two categories: benign and malignant. These labels are essential for training and evaluating our models' performance in accurately diagnosing skin diseases.

To ensure a fair and balanced evaluation, we partitioned the dataset into training and testing sets in an 80:20 ratio. We took care to maintain an equitable distribution of benign and malignant labels within both the training and testing subsets. Additionally, we stratified the data over skin tones, ensuring that each set (training and testing) includes samples from all three skin tone categories shown in Table \ref{tab:train_test_samples}. This approach prevents bias and guarantees that our evaluation dataset is representative of the entire spectrum of skin tones encountered in clinical practice.

\begin{table}[h]
\centering
\caption{Train Test Samples}
\begin{tabular}{ccc}
\hline
\textbf{Samples} & \textbf{Benign} & \textbf{Malignant} \\ \hline
Train (DDI only) & 290 & 103 \\ 
Train (DDI + Ham10000) & 8027 & 1408 \\ 
Val (DDI only) & 97 & 34 \\ 
Test (DDI only) & 98 & 34 \\ \hline
\end{tabular}

\label{tab:train_test_samples}
\end{table}

Furthermore, as part of our preprocessing pipeline, we resized all images to a standard size of 224 pixels and performed common preprocessing techniques to enhance model performance. These preprocessing steps ensure uniformity and facilitate effective model training and testing.

By adhering to these rigorous data processing procedures, we established a robust evaluation framework that enables us to assess the performance of various pretrained models accurately. This approach not only enhances the reliability and reproducibility of our findings but also ensures the inclusivity and fairness of our analysis across diverse skin tone populations.
\subsection{Domain Adaptation}
As we utilize pre-trained models, it's important to note that these models are originally trained on datasets from different domains. For instance, RETFound is trained on retinal datasets. Another reason for domain adaptation is the presence of a small dataset for fine-tuning. Our diverse skin dataset is relatively small. Training or fine-tuning with such a small dataset poses its own challenges. It might lead to suboptimal results. Alternatively, if the model has large weights, it may overfit, resulting in low validation accuracy. This motivates us to add domain adaptation as a prerequisite step for fine-tuning with the DDI dataset. In the domain adaptation step, we use the HAM10000 image dataset, which consists of 10,015 skin images. Our expectation is that domain adaptation will increase the accuracy compared to fine-tuning with DDI only across all benchmarks.

\subsection{Fine-tuning}
\begin{figure}[h!] % h! forces the figure to be placed here
\centering
\includegraphics[width=0.5\textwidth]{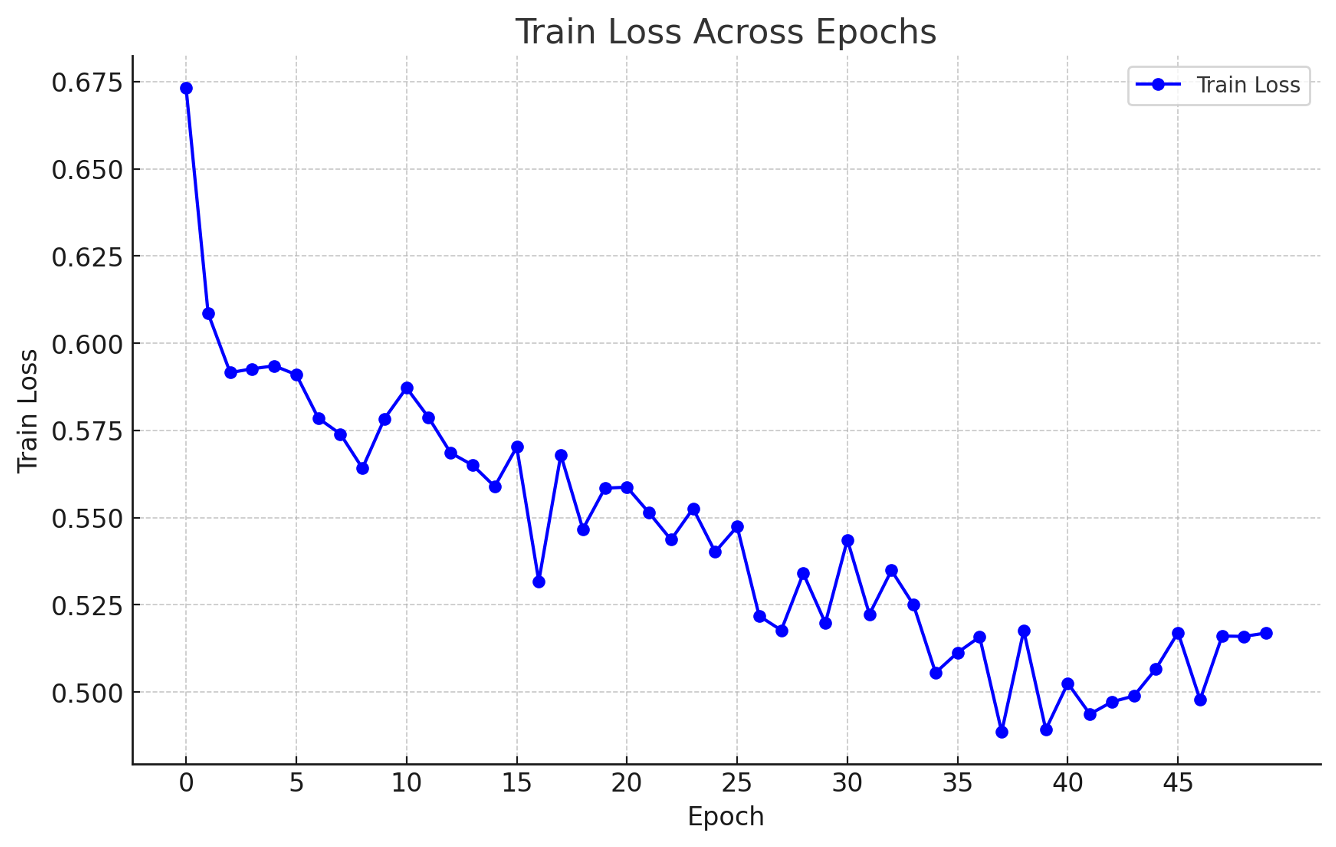} % Adjust the size of the image
\caption{Training loss curve shown for 50 epochs (The training continues up to 100 epochs) during fine-tuning RETFound model on the DDI dataset. }
\label{fig:trainloss}
\end{figure}

Fine-tuning involves adapting a pre-trained model to improve its performance on a specific task by training it further on a task-specific dataset. The tasks utilizing the pre-trained model, RETFound as an example are shown in Figure \ref{fig:model}. For optimal performance on the DDI dataset, we do fine-tuning, which is both effective and resource efficient. We create two datasets - one including HAM10000 samples and another excluding them. Each data set is used to fine-tune our candidate models to explore the hypothesis that domain adaptation enhances performance. In the DDI-only dataset, we include 290 benign and 103 malignant samples for training. The validation dataset comprises 97 benign and 34 malignant samples, while the testing dataset includes 98 benign and 34 malignant samples. Conversely, the dataset that includes HAM10000 contains 8027 benign and 1408 malignant samples. We utilize the Adam optimizer for training over 100 epochs. Referring to Figure~\ref{fig:trainloss}, the training loss plot shows a steep initial decline, indicating rapid learning in the early epochs. The loss then gradually stabilizes with minor fluctuations, which shows the model's convergence.

% ------------ table 2 ---------------

\begin{table*}[ht]
\centering
\caption{Model performance before domain adaptation on the DDI dataset.}
\label{tab:before_adaptation}
\begin{tabular}{ll|cccc}
\hline
\multirow{2}{*}{\textbf{Pre-training Domain}} & \multirow{2}{*}{\textbf{Model}} & \multirow{2}{*}{\textbf{Dataset}} & \multirow{2}{*}{\textbf{Accuracy}} & \multicolumn{2}{c}{\textbf{F1-score}} \\ \cline{5-6}
 & & & & (Macro Avg.) & (Weighted Avg.) \\
\hline
Same domain (Skin) & DeepDerm & DDI & 0.59 & 0.48 & 0.59 \\
                   & HAM10000 & DDI & 0.74 & 0.43 & 0.63 \\
Other Medical Domain & RETFound & DDI & 0.71 & 0.34 & 0.57 \\
                     & YOLOv8-Chest & DDI & 0.69 & 0.54 & 0.67 \\
General Images      & YOLOv8x & DDI & 0.70 & 0.41 & 0.61 \\
                    & YOLOv8n & DDI & 0.73 & 0.61 & 0.71 \\
General Medical Images    & MedViT-small & DDI & 0.70 & 0.57 & 0.70 \\
                    & \textbf{MedViT-base} & \textbf{DDI} & \textbf{0.74} & \textbf{0.60} & \textbf{0.72} \\
                    & MedViT-large & DDI & 0.70 & 0.59 & 0.70 \\
\hline
\end{tabular}
\end{table*}

% ------------ table 3 ---------------

\begin{table*}[ht]
\centering
\caption{Model performance after domain adaptation on combined Ham10000+DDI datasets.}
\label{tab:after_adaptation}
\begin{tabular}{ll|cccc}
\hline
\multirow{2}{*}{\textbf{Pre-training Domain}} & \multirow{2}{*}{\textbf{Model}} & \multirow{2}{*}{\textbf{Dataset}} & \multirow{2}{*}{\textbf{Accuracy}} & \multicolumn{2}{c}{\textbf{F1-score}} \\ \cline{5-6}
 & & & & (Macro Avg.) & (Weighted Avg.) \\
\hline
Same domain (Skin) & DeepDerm & Ham10000+DDI & 0.68 & 0.56 & 0.65 \\
Other Medical Domain & RETFound & Ham10000+DDI & 0.72 & 0.45 & 0.63 \\
                     & YOLOv8-Chest & Ham10000+DDI & 0.71 & 0.56 & 0.68 \\
General Images      & YOLOv8x & Ham10000+DDI & 0.72 & 0.54 & 0.68 \\
                    & YOLOv8n & Ham10000+DDI & 0.73 & 0.56 & 0.69 \\
General Medical Images     & MedViT-small & Ham10000+DDI & 0.74 & 0.63 & 0.73 \\
                    & \textbf{MedViT-base} & \textbf{Ham10000+DDI} & \textbf{0.76} & \textbf{0.63} & \textbf{0.73} \\
                    & MedViT-large & Ham10000+DDI & 0.75 & 0.62 & 0.72 \\
\hline
\end{tabular}
\end{table*}

% -------------- end table ---------------

\subsection{Hyperparameter Selection}
Selecting appropriate hyperparameters is crucial for optimizing the performance of our fine-tuned models. We experimented with various configurations and settled on the following parameters based on their impact on model convergence and accuracy:

\begin{itemize}
    \item \textbf{Batch size}: We use a batch size of 16, which provides a balance between training speed and model stability.
    \item \textbf{Base learning rate (blr)}: The initial learning rate is set to \(5 \times 10^{-3}\). This rate is chosen to ensure fast convergence without overshooting the minima.
    \item \textbf{Layer decay}: A layer decay of 0.65 is applied to adjust the learning rates of deeper layers, effectively preventing overfitting.
    \item \textbf{Weight decay}: We apply a weight decay of 0.05 to regularize the model and reduce the likelihood of overfitting.
    \item \textbf{Drop path rate}: A drop path rate of 0.2 is utilized to introduce regularization by randomly dropping paths during training. This enhance the model's generalization.
    \item \textbf{Number of classes}: Our models are configured to distinguish between two classes, benign and malignant.
    \item \textbf{Input size}: The input size for our models is set to \(224 \times 224 \times 3\), aligning with common practice for image-based models to capture sufficient detail while managing computational load.
\end{itemize}

These hyperparameters were fine-tuned through iterative training and validation, leading to optimized performance on both the training and testing datasets.

% \section{Evaluation}
\subsection{Evaluation metrics}
In our study, we used accuracy, macro-average F1 score and weighted average F1 score to evaluate the performance of various models. Accuracy measures the overall correctness of the model and is defined as the ratio of true predictions (both true positives and true negatives) to the total number of cases examined. The formula for Accuracy is:
\[
\text{Accuracy} = \frac{\text{Number of correct predictions}}{\text{Total number of predictions}}
\]

This metric is straightforward, but may not always provide a complete picture, especially in imbalanced datasets where one class may dominate the others. The F1 score is a harmonic mean of precision and recall, providing a balance between these two metrics. It is particularly useful when dealing with imbalanced datasets. The formula for F1-Score is:
\[
\text{F1-Score} = 2 \times \frac{\text{Precision} \times \text{Recall}}{\text{Precision} + \text{Recall}}
\]

The macro-average method calculates the F1 score independently for each class but does not take class imbalance into account. Each class is given equal weight. The formula for Macro-average F1-Score is:
\[
\text{Macro-average F1} = \frac{\sum(\text{F1-Score of each class})}{\text{Number of classes}}
\]

The weighted-average F1 score calculates the F1 score for each class but gives them a weight depending on their support. This method accounts for class imbalance by weighting the F1-score of each class by the number of true instances in each class. The formula for the weighted average F1 score is:
\begin{multline*}
\text{Weighted-average F1} = \sum \left(\frac{\text{Support of class}}{\text{Total samples}} \times \right. \\
\left. \text{F1-Score of class}\right)
\end{multline*}

Using both macro-average and weighted-average F1 scores and accuracy enables a comprehensive and nuanced evaluation of model performance in classifying skin disease images. The macro-average F1-score highlights the model's consistency across different conditions, emphasizing its capability to handle rare diseases effectively. In contrast, the weighted average F1 score provides insight into the model's accuracy in diagnosing more common diseases, reflecting its practical utility in a typical clinical environment. This layered approach ensures that the evaluation captures both overall accuracy and detailed performance across various class distributions, facilitating a balanced comparison of models tailored to the specific needs of healthcare applications. These metrics provide a comprehensive view of model performance, highlighting strengths in handling the overall dataset and specific classes, particularly useful when dealing with medical data like skin diseases where some conditions may be rarer than others. The evaluation results are presented in Table \ref{tab:before_adaptation} and Table \ref{tab:after_adaptation}. 

\section{Results and Discussions}
\subsection{Comparative Performance Evaluation of Models Before Domain Adaptation Using the DDI Dataset}
Table \ref{tab:before_adaptation} illustrates the initial performance metrics of various models fine-tuned solely on the DDI dataset without domain adaptation. The results show a noticeable variance in performance across different models. Generally, models pretrained on larger, diverse datasets demonstrated superior performance compared to those trained specifically on skin image datasets. For instance, both DeepDerm and HAM10000 exhibited subpar F1 scores, with HAM10000 reaching a relatively high accuracy of 0.74, whereas DeepDerm lagged with an accuracy of 0.59. RETFound and YOLOv8 outperformed these models, showing better accuracy and F1 scores. Specifically, YOLOv8 models trained on a broader range of data outperformed those trained exclusively on chest images or the YOLOv8-x variant.

The enhanced adaptability of YOLOvN to skin images might come from its more effective feature extraction and augmentation techniques, which are crucial to handling nuanced variations in skin texture and color. This model also appears to better generalize across the smaller, more specialized datasets typical in dermatology, potentially reducing overfitting issues seen in YOLOv8-x.

Among all models evaluated, MedVIT stood out, probably due to its medical imagery-optimized transformer architecture that can ably integrate multiscale features and fine-grained details essential for accurate skin condition classification. This design is particularly adept at utilizing sparse annotations prevalent in medical datasets, thereby boosting its learning efficiency. Moreover, within the MedVIT series, the base model distinguished itself by achieving the highest accuracy at 0.74, surpassing both the small and large versions. This superior performance is attributed to its balanced complexity, which effectively prevents overfitting, and its focused and efficient feature learning capabilities.

\subsection{Enhanced Model Performance Through Domain Adaptation}
As detailed in Table \ref{tab:after_adaptation}, post-domain adaptation—which involved fine-tuning models on a combined dataset of HAM10000 and DDI—significant improvements were noted in both accuracy and F1 scores across most models. This process leveraged the larger HAM10000 dataset, which comprises approximately 10,000 samples, significantly more than the DDI dataset. This considerable dataset size helped bridge the domain gap and enhanced the models' ability to capture and learn from diverse skin image features more effectively.

However, the performance of YOLOv8-N slightly decreased, likely due to the model overfitting when exposed to the large volume of domain-specific data. In contrast, other models demonstrated substantial enhancements due to domain adaptation: DeepDerm showed notable increases of 15\% in accuracy and 16.7\% in macro-average F1 score, indicating that additional training on skin images significantly bolstered its capability for efficient feature learning. RETFound also displayed improved performance, with gains of 1.4\% in accuracy and 4.7\% in macro-average F1 score.

Furthermore, MedViT-base saw its accuracy rise from 0.74 to 0.76, along with improvements of 2.7\% in accuracy and 5\% in macro-average F1 score. These results underscore that incorporating more domain-specific training data can significantly enhance model robustness and accuracy, making them more adept at classifying various skin diseases.

\begin{figure*}[t]
\centering
\includegraphics[width=0.95\textwidth]{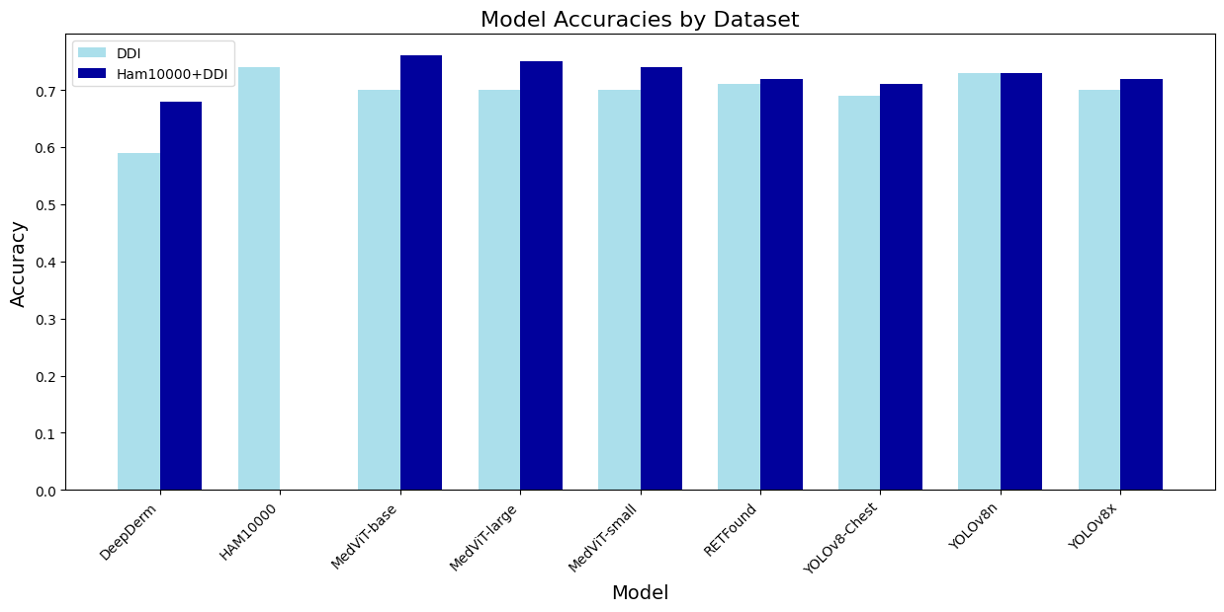} % Reduce the 
\caption{Comparative analysis of model performances across DDI and Ham10000+DDI datasets, highlighting the efficacy of domain adaptation and model scaling. The comparison includes a range of models: DeepDerm and Ham10000, specifically trained on skin disease images; MedViT, a model pretrained on general medical images; YOLOv8, adapted from general imaging tasks; and RETFOUND, originating from retinal image analyses. Notably, MedViT-base outperforms other models in adapting to both datasets, showcasing its robustness and versatility in domain adaptation scenarios.}
\label{fig:evaluation_results}
\end{figure*}

% \begin{figure}[t] % h! forces the figure to be placed here
% \centering
% \includegraphics[width=0.5\textwidth]{evaluation_results.png} % Adjust the size of the image
% \caption{}
% \label{fig:evaluation_results}
% \end{figure}

\subsection{Discussions}
As shown in Fig~\ref{fig:evaluation_results}, the results of our experiments prove the effectiveness of transfer learning approach. First, the improvement in model performance due to domain adaptation is evident from the comparison between the DDI-only models and those fine-tuned on combined Ham10000+DDI datasets. Models pre-trained on skin images, such as DeepDerm and HAM10000, demonstrate a notable increase in accuracy when further adapted to specific dermatological tasks. This underscores the benefit of using domain-specific training data, which enhances the model's ability to generalize from learned dermatological features. Secondly, general medical image models, such as MedViT, often outperform domain-specific models. This can be attributed as their training on diverse medical imagery, enabling them to learn more robust and generalizable features. MedViT-base achieves higher accuracy compared to more specialized models like DeepDerm and RETFound. The data also reveals that larger versions of models, such as MedViT-large, do not always equate to better performance. In some instances, these models exhibit a decline in accuracy compared to their base or smaller counterparts, likely due to overfitting on the training data. The larger models, while potentially more powerful, might be too complex for the amount of training data available, leading to worse generalization on unseen data.

\section{Conclusion}
In this article, we showed a study that represents a significant advancement in the field of dermatological AI, addressing critical challenges in the diagnosis of skin diseases and paving the way for more inclusive and accurate diagnostic tools. Through comprehensive experimentation and analysis, we have demonstrated the effectiveness of pre-trained models, including RETFound, MedViT, and YOLOv8-Chest, in accurately predicting skin diseases in various skin tones. Our research underscores the importance of leveraging transfer learning techniques and domain adaptation to harness the wealth of knowledge encapsulated in pre-trained models, thereby enhancing their performance on underrepresented skin tones. By benchmarking these models on the Diverse Dermatology Images (DDI) dataset, we have provided valuable insights into their strengths and limitations, allowing clinicians and researchers to make informed decisions regarding model selection and deployment. Furthermore, our meticulous data processing procedures, including stratification by skin tone and label balance, ensure the fairness and reliability of our evaluation framework. This approach not only improves the generalizability of our findings, but also underscores our commitment to inclusivity and equity in dermatological AI research. Looking ahead, our findings lay the foundation for future research efforts aimed at further improving the accuracy and inclusiveness of skin disease diagnosis. By continuing to refine and expand upon our methodologies, we can drive innovation in the development of AI-driven diagnostic tools that benefit patients of all skin tones.

\section{Future works}
In our current research, we've utilized pre-trained models from various domains and perspectives to enhance the performance of our model. For instance, we have leveraged pre-trained models designed for analyzing retinal images, chest images, and medical images such as MedVit, which specializes in medical image analysis. Using transfer learning techniques, we adapt these models to work with skin images, broadening the scope of our diagnostic capabilities.

Looking ahead, our future work will involve evaluating the effectiveness of SkinGPT \cite{zhou2023skingpt}, a generative model specifically designed to generate prompts to diagnose and describe skin conditions. We plan to evaluate SkinGPT's performance using the diverse DDI dataset, which will allow us to gauge its ability to handle a wide range of skin images for diagnostic tasks. This evaluation step is crucial for understanding the model's strengths and limitations in real-world applications.

Furthermore, we intend to incorporate other state-of-the-art methods for skin image prediction into our performance benchmarking process. By including these alternative models, we aim to provide a comprehensive comparison of different approaches in the field, offering valuable insights into their respective capabilities and performance metrics.

In addition to evaluating the diagnostic accuracy of these models, we will also consider the computational efficiency of each approach. Comparing the computation time required by different models will serve as an essential benchmarking metric, helping to inform decisions regarding model selection and deployment in practical settings.

Overall, our future research endeavors aim to advance the state-of-the-art in skin image analysis by evaluating the performance of novel generative models like SkinGPT and conducting comprehensive benchmarking analyses to guide the development of more effective and efficient diagnostic tools.

% \subsubsection{Subsubsection Heading Here}
% Subsubsection text here.

% conference papers do not normally have an appendix

% % use section* for acknowledgement
% \section*{Acknowledgment}

% The authors would like to thank Virginia Tech 

% trigger a \newpage just before the given reference
% number - used to balance the columns on the last page
% adjust value as needed - may need to be readjusted if
% the document is modified later
%\IEEEtriggeratref{8}
% The "triggered" command can be changed if desired:
%\IEEEtriggercmd{\enlargethispage{-5in}}

% references section

% can use a bibliography generated by BibTeX as a .bbl file
% BibTeX documentation can be easily obtained at:
% http://www.ctan.org/tex-archive/biblio/bibtex/contrib/doc/
% The IEEEtran BibTeX style support page is at:
% http://www.michaelshell.org/tex/ieeetran/bibtex/
% \bibliographystyle{IEEEtran}
% argument is your BibTeX string definitions and bibliography database(s)
%\bibliography{IEEEabrv,../bib/paper}
%
% <OR> manually copy in the resultant .bbl file
% set second argument of \begin to the number of references
% (used to reserve space for the reference number labels box)
% \begin{thebibliography}{1}

% \bibitem{IEEEhowto:kopka}
% H.~Kopka and P.~W. Daly, \emph{A Guide to \LaTeX}, 3rd~ed.\hskip 1em plus
%   0.5em minus 0.4em\relax Harlow, England: Addison-Wesley, 1999.

% \end{thebibliography}

% \bibliographystyle{plain} % Defines the bibliography style
\bibliography{CameraReady/LaTeX/reference} % The filename of the .bib file, without the extension
% that's all folks
\end{document}